\documentclass[10pt,twocolumn,letterpaper]{article}

\usepackage{cvpr}
\usepackage{times}
\usepackage{epsfig}
\usepackage{graphicx}
\usepackage{amsmath}
\usepackage{amssymb}

\usepackage[breaklinks=true,bookmarks=false]{hyperref}

\cvprfinalcopy 

\def\cvprPaperID{****} 

\begin{document}

\title{Neural Aesthetic Image Reviewer}

%
%
\author{Wenshan Wang$^1$, Su Yang$^{1,3}$\thanks{Correspondence author}, Weishan Zhang$^2$, Jiulong Zhang$^3$ \\$^1$Shanghai Key Laboratory of Intelligent Information Processing \\ School of Computer Science, Fudan University\\ $^2$China University of Petroleum \\ $^3$Xi'an University of Technology \\ {\tt\small \{wswang14, suyang\}@fudan.edu.cn}
\
}

\maketitle

\begin{abstract}
  Recently, there is a rising interest in perceiving image aesthetics. The existing works deal with image aesthetics as a classification or regression problem. To extend the cognition from rating to reasoning, a deeper understanding of aesthetics should be based on revealing why a high- or low-aesthetic score should be assigned to an image. From such a point of view, we propose a model referred to as Neural Aesthetic Image Reviewer, which can not only give an aesthetic score for an image, but also generate a textual description explaining why the image leads to a plausible rating score. Specifically, we propose two multi-task architectures based on shared aesthetically semantic layers and task-specific embedding layers at a high level for performance improvement on different tasks. To facilitate researches on this problem, we collect the AVA-Reviews dataset, which contains 52,118 images and 312,708 comments in total. Through multi-task learning, the proposed models can rate aesthetic images as well as produce comments in an end-to-end manner. It is confirmed that the proposed models outperform the baselines according to the performance evaluation on the AVA-Reviews dataset. Moreover, we demonstrate experimentally that our model can generate textual reviews related to aesthetics, which are consistent with human perception.
\end{abstract}

\section{Introduction}

\noindent The problem of perceiving image aesthetics, styles, and qualities has been attracting increasingly much attention recently. The goal is to train an artificial intelligence (AI) system being able to perceive aesthetics as human. In the context of computer vision, automatically perceiving image aesthetics has many applications, such as personal album management systems, picture editing software, and content based image retrieval systems.

 \begin{figure}[ht]
  \begin{minipage}{0.49\linewidth}
    \center
    \includegraphics[width=1.0\linewidth]{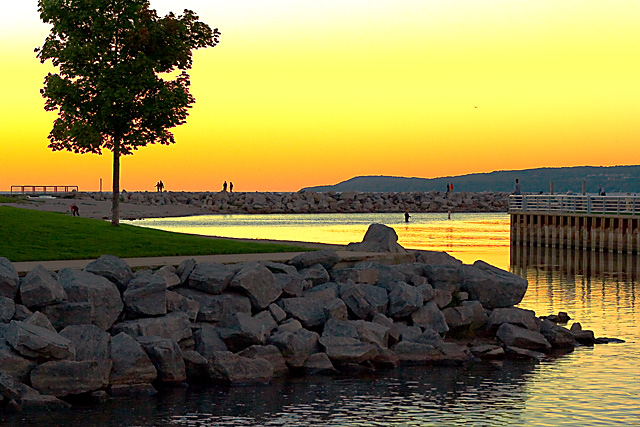}
  \end{minipage}%
  \hfill
  \begin{minipage}{0.49\linewidth}
   \textcircled{\small{1}}\textbf{Prediction}:High-aesthetic category\\
   \textcircled{\small{2}}\textbf{Comments}:Fantastic colors and great sharpness.
  \end{minipage}
  \vfill
  \begin{minipage}{0.48\linewidth}
   \textcircled{\small{1}}\textbf{Prediction}:Low-aesthetic category\\
   \textcircled{\small{2}}\textbf{Comments}:The focus seems a little soft.
  \end{minipage}
  \hfill
  \begin{minipage}{0.51\linewidth}
    \center
    \includegraphics[width=1.0\linewidth]{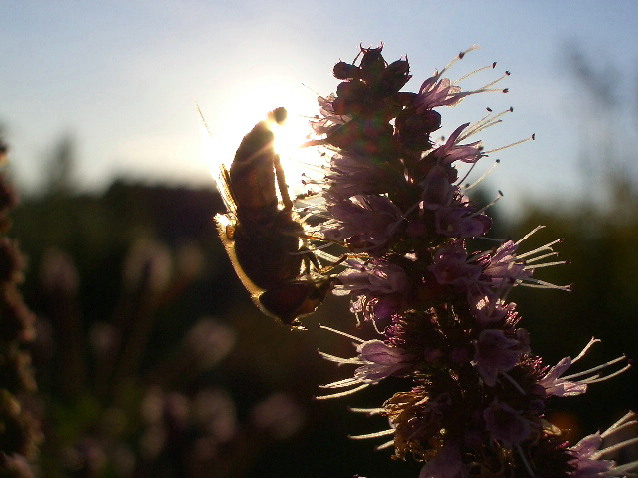}
  \end{minipage}
  \caption{Some examples to clarify the problem: The goal is to perceive image aesthetics
as well as generate reviews or comments.}
  \label{fig:problem_setting}
\end{figure}

 In the literature, the problem of perceiving image aesthetics is formulated as a classification or regression problem. However, human vision and intelligence can not only perceive image aesthetics, but also generate reviews or explanations, which express aesthetic insights in a natural language way. Motivated by this, we propose a new task of automatically generating aesthetic reviews or explanations. Imagine such a scenario that a mobile phone user takes a photo and upload it to an AI system, the AI system can automatically predict a score indicating high or low rating in the sense of aesthetics and generate reviews or comments, which express aesthetic insights or give advices on how to take photos being consistent with aesthetic principles.



 Formally, we consider the novel problem as two tasks: One is assessment on image aesthetics and the other is generating textual descriptions related to aesthetics, which is important in that it enables understanding image aesthetics not only from the phenomenon point of view but also from the perspective of mechanisms. In Fig. \ref{fig:problem_setting}, we demonstrate the rising of the problem by a couple of examples. The problem that we explore here is to predict the images of interest as high- or low-aesthetic based on the visual features while producing the corresponding comments. For instance, as shown in Fig. \ref{fig:problem_setting}, our model can not only perform prediction on image aesthetics, but also produce the comments ``Fantastic colors and great sharpness'' for the top-left image. In contrast to the image aesthetics classification task \cite{7974874,Lu2015RAPID,Lu2015Deep}, we extend it by adding a language generator, which can produce aesthetic reviews or insights, as being helpful to understand aesthetics. Unlike image captioning \cite{7298935,7780863,7780398}, for which the existing works are focused on producing factual descriptions while neglecting generation of descriptions of styles, we explore the problem of generating descriptions in terms of aesthetics.

 To facilitate this research, we collect the AVA-Reviews dataset from Dpchallenge, which contains 52,118 images and 312,708 comments in total. On the Dpchallenge website, users can not only rate images with an aesthetic score for each image, but also give reviews or explanations that express aesthetic insights. For instance, comments such as ``an amazing composition'' express high-aesthetic while comments like ``too noise'' represent low-aesthetic.  Motivated by this observation, we use the users' comments from Dpchallenge as our training corpus.

  \begin{figure}[t]
    \centering
    \includegraphics[width=0.48\textwidth]{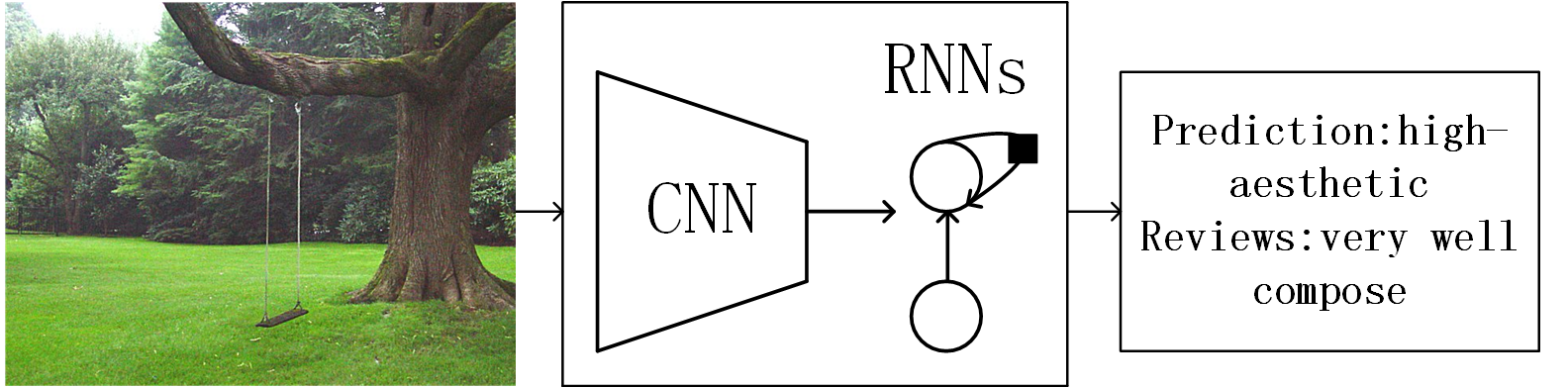}
    \caption{Our multi-task framework based on CNN plus RNNs, which consists of an aesthetic image classifier and a language generator.}
    \label{fig:intro_framework}
 \end{figure}


 In this paper, we propose a novel model referred to as Neural Aesthetic Image Reviewer (NAIR), which can not only distinguish high- or low-aesthetic images but aslo automatically generate aesthetic comments based on the corresponding visual features of such images. Motivated by \cite{7298935}, we construct NAIR model by adopting the convolution neural network (CNN) plus Recurrent Neural Networks (RNNs) framework, which encodes any given image into a vector of fixed dimension, and decodes it to the target description. As shown in Fig. \ref{fig:intro_framework}, our framework consists of an aesthetic image classifier based on CNN, which can predict images as high- or low-aesthetic, and a language generator based on RNNs, which can produce reviews or comments based on high-level CNN features.
 Specifically, we propose two novel architectures based on shared aesthetically semantic layers and task-specific embedding layers at a high level
 to improve performance of different tasks through multi-task learning. Here, the motivation to propose the models as such is that it is a common practice to make use of shared layers at a high level in neural networks for performance improvement on different tasks. We evaluate the NAIR model on the AVA-Reviews dataset. The experimental results show that the proposed models can improve the performance on different tasks compared to the baselines. Moreover, we demonstrate that our model can produce textual comments consistent with human intuition on aesthetics. Our contributions are summarized as follows:

 \begin{itemize}
  \item 
  The problem that we explore here is whether computer vision systems can perceive image aesthetics as well as generate reviews or explanations as human. To the best of our knowledge, it is the first work to investigate into this problem.
  \item By incorporating shared aesthetically semantic layers at a high level, we propose an end-to-end trainable NAIR architecture, which can approach the goal of performing aesthetic prediction as well as generating natural-language comments related to aesthetics.
  \item To enable this research, we collect the AVA-Reviews dataset, which contains 52,118 images and 312,708 comments. We hope that this dataset could promote researches on vision-language based image aesthetics.
 \end{itemize}

\section{Related Work}

 In this section, we review two related topics:

 One topic is deep features for aesthetic classification. The typical pipeline for image aesthetics classification is to classify the aesthetic image of interest as high- or low-aesthetic in the context of supervised learning. Since the work by \cite{Krizhevsky2012ImageNet}, the powerful deep features have shown a good performance on various computer vision tasks. That inspires the use of deep features to improve the classification performance for image aesthetics. Below, we review the recent works based on deep CNN features. The existing approaches \cite{Kang2014Convolutional,Lu2015RAPID,Lu2015Deep} aim to improve the performance of classification by extracting both global- and local-view patches from high-resolution aesthetic images. Yet, the improved classification accuracy does not make sense in revealing the underlying mechanism of human or machine perception in terms of image aesthetics. In view of such limit of the existing researches, we extend the traditional aesthetic image assessment task by adding a language generator, which is helpful to understand the mechanism of image aesthetics as well as reason the focus of the visual perception of people from the aesthetic perspective.

 The other topic is generation of vision-to-language descriptions. Generating image descriptions with texts automatically is an important problem in artificial intelligence. The recent works take advantage of deep learning to generate natural-language descriptions to describe image contents because deep learning based methods in general promise superior performance. A typical pipeline is  the encoder-decoder based architecture, which is transfered from neural machine translation \cite{NIPS2014_5346} to computer vision. Modelling the probability distribution in the space of visual features and textural sentences leads to generating more novel sentences. For instance, in order to generate textural descriptions for an image, \cite{7298935} proposes an end-to-end CNN-LSTM based architecture, where CNN feature is considered as a signal to start LSTM. \cite{icml2015_xuc15} improves the image captioning performance using attention mechanism, where the different regions of the image can be selectively attended when generating a word in a step. \cite{7780398,7780872} achieves a significant improvement on image captioning with high-level concepts/attributes predicted by CNN. \cite{7780863} and \cite{Krishna2017} employ the rich information of individual regions in images to generate dense image captions.

 In the literature, despite the progress made in image captioning, the most existing approaches only generate factual descriptive sentences. Notably, the trend of the recent researches has been shifted to produce non-factual descriptions. For example, \cite{Mathews:2016:SGI:3016387.3016406} generates image captions with sentiments, and \cite{Gan_2017_CVPR} proposes the StyleNet to produce humorous and romantic captions. However, the generation of image descriptions related to art and aesthetics remains an open problem yet. In this work, we propose an end-to-end multi-task framework that can not only classify aesthetic images, but also generate sentences in terms of aesthetics for images.

\section{Neural Models for Image Aesthetic Classification and Vision-to-Language Generation}

 In this section, we introduce the deep learning architectures for image aesthetic classification and vision-to-language generation. Since deep neural models achieve superior performance in various tasks, deep CNN \cite{Lu2015RAPID,Lu2015Deep} is used for image aesthetic classification. Further, CNN plus RNNs architecture \cite{7298935} is used for generating vision-to-language descriptions.

\subsection{Image Aesthetic Classification}
\label{cnn}
 Here, we adopt the single-column CNN framework \cite{Lu2015RAPID} for image aesthetic classification. For the task of binary classification, the learning process of CNN is as follows. Given a set of training examples $\left\{ {\left( {{x_i},{y_i}} \right)} \right\}$, where ${x_i}$ is the high-resolution aesthetic image, and ${y_i} \in \left\{ {0,1} \right\}$ the aesthetic label, then we minimize the cross-entropy loss defined as follows:
 \begin{equation}
 \label{loss_aesthetics}
\begin{split}
 {L_{aesthetics}}\left( \theta  \right) =  - \frac{1}{n}\sum\limits_{i = 1}^n {\sum\limits_{y_i} {\{ y_i\log p\left( {{{\hat y}_i} = y_i|{x_i};\theta } \right)} } \\
 + \left( {1 - y_i} \right)\log \left( {1 - p\left( {{{\hat y}_i} = y_i|{x_i};\theta } \right)} \right)\}
\end{split}
 \end{equation}
 where $p\left( {{{\hat y}_i} = y_i|{x_i};\theta } \right)$ is the probability output of the softmax layer, and $\theta$ the weight set of CNN.


\subsection{Vision-to-Language Generation}

   \begin{figure}[t]
    \centering
    \includegraphics[width=0.45\textwidth]{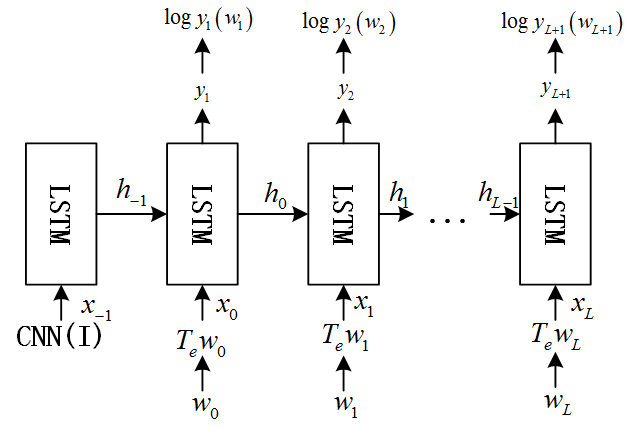}
    \caption{Unrolled LSTM model along with an image representation based on CNN and word embedding.}
    \label{fig:lstm}
 \end{figure}

 \begin{figure*}[ht]
    \centering
    \includegraphics[width=0.95\textwidth]{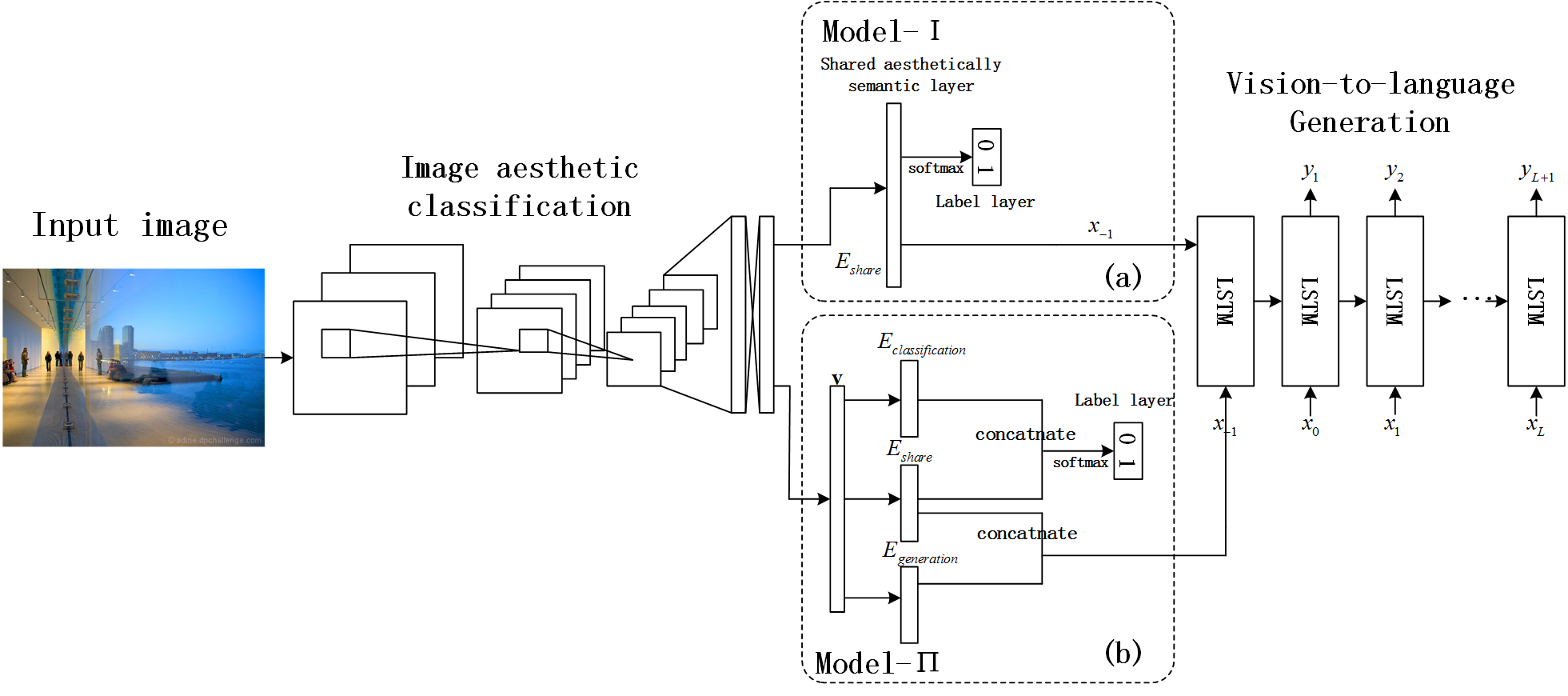}
    \caption{The proposed models based on CNN plus RNNs architecture. It consists of an image aesthetic classification part, which performs binary classification using the single-column CNN, and a vision-to-language generation part, which generates natural-language comments using pipelined LSTM models. Model-\uppercase\expandafter{\romannumeral1} includes the shared aesthetically semantic layer. Model-\uppercase\expandafter{\romannumeral2} includes the task-specific embedding layer.}
    \label{framework}
\end{figure*}

 Following the previous works \cite{NIPS2014_5346,7298935,7558228,DBLP:journals/corr/YaoPLQM16}, we adopt the architecture of CNN plus RNNs to generate textural descriptions for images. The key of these approaches is to encode a given image into a fixed-dimension vector using CNN, and then decode it to the target output description. Formally, suppose that a training example pair $\left( {S,I} \right)$ is given, where $I$ and $S = \left\{ {{w_1},{w_2},...,{w_L}} \right\}$ denote an image and a textual description, respectively, and $L$ is the length of the description. The goal of the description generation is to minimize the following loss function:
 \begin{equation}
  \label{loss_language}
  {L_{language}}\left( {I,S} \right) =  - \log p\left( {S|I} \right)
 \end{equation}
 where $\log p\left( {S|I} \right)$ is the log probability of the correct description given the visual features $I$. Since the model generates words in the target sentence one by one, the chain rule can be applied to learn the joint probability over the context words. Thus, the log probability of the description can be given by the sum of the log probabilities over the words as
 \begin{equation}
  \label{sum_log_pro}
  \log p\left( {S|I} \right) = \sum\limits_{t = 1}^L {\log p\left( {{w_t}|I,{w_0},{w_1},...,{w_{t - 1}}} \right)}
 \end{equation}
 At the training stage, we minimize the loss as shown in (\ref{sum_log_pro}) to guarantee the contextual relationship among words.

 Naturally, we model the log probability as described in (\ref{sum_log_pro}) with Long Short-Term Memory (LSTM), which is a variant of RNN, capable of learning long-term dependencies \cite{Hochreiter:1997:LSM:1246443.1246450}. We can train the LSTM model to generate descriptions for images in an unrolled form as shown in Fig. \ref{fig:lstm}. First, we feed the LSTM model $L+2$ words $S = \left\{ {{w_0},{w_1},...,{w_L},{w_{L + 1}}} \right\}$, where ${w_0}$, ${w_{L + 1}}$ represent a special START token "$</{\rm{S}}>$" and a special END token "$< /{\rm{E}} >$" of the description. At time step $t =  - 1$, we set ${x_{ - 1}} = {\rm{CNN}}\left( I \right)$, where an input image $I$ is represented by CNN. From time $t = 0$ to $t = L$, we set ${x_t} = {T_e}{w_t}$, and then LSTM computes the hidden state ${h_t}$ and the output probability vector ${y_t}$ using a recurrence formula $\left( {{h_t},{y_{t + 1}}} \right) = {\rm{LSTM}}\left( {{h_{t - 1}},{x_t}} \right)$, where $T_e$ denote the weights of word embedding \cite{DBLP:journals/corr/abs-1301-3781}.

 Following \cite{7298935}, we feed the image representation ${x_{ - 1}}$ to the LSTM model as input, and then use BeamSearch to generate a description at the testing stage.  

\section{Neural Aesthetic Image Reviewer}


 \textbf{Overview}: Our model aims to predict high- or low-aesthetic categories of images while automatically generate natural-language comments related to aesthetics. We solve this as a multi-task problem in an end-to-end deep learning manner. Similar to \cite{7298935}, we adopt the CNN plus RNNs architecture, as illustrated in Fig. \ref{framework}. The architecture consists of an image aesthetic classification part and a vision-to-language generation part. In the image aesthetic classification part, we use the single-column CNN framework for the task of binary classification. Further, the CNN framework can produce a high-level visual feature vector for an image, which can be fed to the vision-to-language part. In the vision-to-language part, we use LSTM based models to generate natural-language comments. We feed the high-level visual feature vector as an input to pipelined LSTM models. During inference, the procedure of generating textural descriptions on visual scenes is illustrated in Fig. \ref{fig:lstm}. When training the multi-task model, we feed a training instance to the model repeatedly, which consists of an image, the corresponding aesthetic category label, and a ground-truth comment. Given a test image, the model automatically predicts the image of interest as high- or low-aesthetic while outputs a comment.

 By utilizing multi-task learning, we propose three neural models based on the CNN plus RNNs architecture to approach the goal of performing aesthetic prediction as well as generating natural-language comments related to aesthetics. Below, we describe the details of the proposed models.

 \textbf{Multi-task baseline}: Here, we propose a baseline multi-task framework. One natural way is to directly sum up the loss for image aesthetic classification ${L_{aesthetics}}$ and the loss for textural description generation ${L_{language}}$, which is formulated as
 \begin{equation}
  \label{fun_joint}
  {L_{joint}} = \alpha {L_{aesthetics}} + \beta {L_{language}}
 \end{equation}
 where ${L_{joint}}$ is the joint loss for both tasks, and $\alpha$, $\beta$ control the relative importance of the image aesthetic classification task and the language generation task, respectively, which are set based on validation data.

 In this model, we minimize the joint loss function ${L_{joint}}$ by changing the weights of the CNN components and the RNNs components at the same time. Through the experiment, we find that changing the weights of CNN components has a negative effect.

 \textbf{Model-\uppercase\expandafter{\romannumeral1}}: Motivated by the Multi-task baseline, we minimize the joint loss function ${L_{joint}}$ by fixing the weights of CNN components while introducing a shared aesthetically semantic layer, allowing two different tasks to share information at a high level, which is illustrated in Fig. \ref{framework}(a). Suppose that we have visual feature vector ${\bf{v}}$ as image representation. Then, we can formulate the shared aesthetically semantic layer ${E_{share}}$ as
 \begin{equation}
  \label{eq:share}
 {E_{share}} = f\left( {{W_s}{\bf{v}}} \right)
 \end{equation}
 where $f$ can be a non-linear function such as ReLU function, and ${{W_s}}$ is the weights for learning.

 \textbf{Model-\uppercase\expandafter{\romannumeral2}}: The potential limitation of Model-\uppercase\expandafter{\romannumeral1} is that some task-specific features can not be captured by the shared aesthetically semantic layer. To address this problem, we introduce a task-specific embedding layer for each task in addition to the shared aesthetically semantic layer as described in Model-\uppercase\expandafter{\romannumeral1}.

 Formally, we introduce the classification-specific embedding layer ${E_{classification}}$ and the generation-specific embedding layer ${E_{generation}}$ for the image aesthetic classification task and the vision-to-language generation task, respectively, which can be defined as follows:
 \begin{equation}
  {E_{classification}} = f\left( {{W_c}{\bf{v}}} \right)
 \end{equation}
 \begin{equation}
  {E_{generation}} = f\left( {{W_g}{\bf{v}}} \right)
 \end{equation}
 where ${{W_c}}$ and ${{W_g}}$ are learnable parameters. In addition, we have the shared aesthetically semantic layer ${E_{share}}$, which is defined in Equation (\ref{eq:share}).  Then, we can concatenate the task-specific embedding feature vector and the shared aesthetically semantic feature vector as the final feature representation, which is shown in Fig. \ref{framework}(b).

\section{Experiments}
 We conduct a series of experiments to evaluate the effectiveness of the proposed models based on the newly collected AVA-Reviews dataset.

\subsection{AVA-Reviews Dataset}
 The AVA dataset \cite{6247954} is one of the largest datasets for studying image aesthetics, which contains more than 250,000 images download from a social network, namely, Dpchallenge\footnote{http://www.dpchallenge.com/}. Each image has a large number of aesthetic scores ranging from one to ten obtained via crowdsourcing. Further, the Dpchallenge website allows users to rate and comment on images. These reviews express users' insights into why to rate an image as such, and further give guidelines on how to take photos. For instance, users use comments such as ``amazing composition'', ``such fantastic cropping and terrific angle'', or ``very interesting mix of warm and cold colors, good perspective'' to express reasons for giving high rating. In contrast, users use reviews such as ``too much noise'', ``a bit too soft on the focus'', or ``colors seem a little washed out'' to indicate why they give low rating. Based on this observation, we use the users' comments from Dpchallenge as our training corpus.

 Here, we describe how to extract AVA-Reviews from the AVA dataset to compose the AVA-Reviews dataset for evaluation. Following the experimental settings in \cite{6247954}, the images with average scores less than $5 - \delta$ are low-aesthetic images, while the images with average scores greater than or equal to $5 + \delta $ are high-aesthetic images, where $\delta$ is a parameter to discard ambiguous images. In the AVA-Reviews dataset, we let $\delta$ to be 0.5, and then we randomly select examples from high-aesthetic images and low-aesthetic images to form the training set, validation set, and testing set, respectively. Besides, we crawl all the comments from Dpchallenge for each image in the AVA-Reviews dataset. Each image has six comments. The statistics of the AVA-Reviews dataset are shown in Table \ref{dataset}.

\begin{table}[htbp]
  \centering
    \begin{tabular}{|c|ccc|}
        \hline
          & \multicolumn{1}{c}{Train} & \multicolumn{1}{c}{Validation} & \multicolumn{1}{c|}{Test} \\
        \hline
        \hline
    high-aesthetic images & 20,000 & 3,000 & 3,059 \\
    low-aesthetic images & 20,000 & 3,000 & 3,059 \\
    Totol Images & 40,000 & 6,000 & 6,118 \\
    Reviews &   240,000    &    36,000   & 36,708 \\
    \hline
    \end{tabular}%
  \label{dataset}
  \caption{Statistics of the AVA-Reviews dataset}
\end{table}%


\subsection{Parameter Settings and Implementation Details}
 Following \cite{7298935}, we perform basic tokenization on the comments in the AVA-Reviews dataset. We filter out the words that appear less than four times, resulting in the vocabulary of 13400 unique words. Each word is represented as a ``one-hot'' vector\footnote{https://en.wikipedia.org/wiki/One-hot} with the dimension equal to the size of the word dictionary. Then, the ``one-hot'' vector is transformed to 512-dimensional word embedding vector. Besides, we use the 2,048-dimensional vector as image representation, which is output from the Inception-v3 model pre-trained on ImageNet.


 We implement the model using the open-source software TensorFlow \cite{DBLP:journals/corr/AbadiABBCCCDDDG16}. Specifically, we let the number of the LSTM units to be 512. In order to avoid overfitting, we apply the dropout technique \cite{DBLP:journals/corr/abs-1207-0580} to LSTM variables with the keep probability of 0.7. We initialize all the weights with a random uniform distribution except for the weights of the Inception-v3 model. Further, in Model-\uppercase\expandafter{\romannumeral1}, we let the units of the shared aesthetically semantic layer to be 512. In Model-\uppercase\expandafter{\romannumeral2}, we set the units of the shared aesthetically semantic layer and the task-specific embedding layer to be identically 256.

 To train the proposed models, we use stochastic gradient descent with the fixed learning rate 0.1 and the mini-batch size 32. For multi-task training, we can tune the weights $\alpha$, $\beta$ using the validation set to obtain the optimal parameter values. At the testing stage, we use the BeamSearch approach to generate comments with a beam of size 20 when inputing an image.

\begin{table*}[t]
  \centering
    \begin{tabular}{|c|cccccccc|}
    \hline
      Model    & Over accuracy  & BLEU-1 & BLEU-2 & BLEU-3 & BLEU-4 & METEOR & ROUGE & CIDEr \\
    \hline
    \hline
    IAC   &    72.96   & -    &    -   &    -   &    -   &   -    &    -   & - \\
    V2L &   -    &   47.0    &   24.3    &  13.0     &   6.9    &   10.4    &   24.4    & 5.2 \\
    MT baseline  & 70.57 & 45.2 & 23.5 & 12.7 & 5.9 &  9.7   &   24.1    &  4.2\\
    Model-\uppercase\expandafter{\romannumeral1}  & 75.05 &  47.1 & 24.9 & 13.8 & 7.0 & 11.0 & 25.0 & 5.5 \\
    Model-\uppercase\expandafter{\romannumeral2} & 76.46 & 49.5 & 26.4 & 14.5 & 7.4 & 11.5 & 26.1 & 6.0 \\
    \hline
    \end{tabular}%
  \label{tab:evalution}%
  \caption{Performance of the proposed models on the AVA-Reviews dataset. We report overall accuracy , BLEU-1,2,3,4, METEOR, ROUGE, and CIDEr. All values refer to percentage (\%).}
\end{table*}%

\subsection{Evaluation Metrics}
 We evaluate the proposed method in two ways:

 For the image aesthetic assessment model, we report the overall accuracy, which is used in \cite{Luo:2008:PVQ:1478172.1478204}, \cite{Lu2015RAPID}, and \cite{Lu2015Deep}. It can be formulated as
 \begin{equation}
  {\rm{Overall \quad accuracy = }}\frac{{TP + TN}}{{P + N}}
 \end{equation}
 where $TP$, $TN$, $P$, $N$ denote true positive examples, true negative examples, total positive examples, and total negative examples, respectively.

 To evaluate the proposed language generation model, we utilize four metrics including BLEU@N \cite{Papineni:2002:BMA:1073083.1073135}, METEOR \cite{Lavie:2007:MAM:1626355.1626389}, ROUGE \cite{lin:2004:ACLsummarization}, and CIDEr \cite{7299087}. For all the metrics, a larger value means better performance. The evaluation source code\footnote{https://github.com/tylin/coco-caption} is released by Microsoft COCO Evaluation Server \cite{DBLP:journals/corr/ChenFLVGDZ15}.

\subsection{Baselines for Comparsion}
 In order to evaluate the performance of the proposed multi-task models, we compare them with the following models including two single-task models (image aesthetic classification and vision-to-language) and a multi-task baseline.
 \begin{itemize}
  \item \textbf{Image Aesthetic Classification (IAC)}: We implement the single-column CNN framework \cite{Lu2015RAPID} to predict the image of interest as high- or low-aesthetic images. Unlike \cite{Lu2015RAPID}, we use the more powerful Inception-v3 model \cite{7780677} as our architecture, which is pre-trained on ImageNet. Following \cite{7780677}, the size of every input image is fixed to 299 $\times$ 299 $\times$ 3.
  \item \textbf{Vision-to-Language (V2L)}: We implement a language generation model conditioned on the images undergoing Neural Image Caption \cite{7298935}, which is based on encoder-decoder architecture. Here, we train this model on the AVA-Reviews dataset.
  \item \textbf{Multi-task baseline (MT baseline)}: We implement this model by minimizing the loss function as shown in (\ref{fun_joint}). Note that we change the weights of the CNN components and the weights of the RNNs components at the same time.
\end{itemize}


\subsection{Experimental Results}

 Table \ref{tab:evalution} reports the performance of the proposed models based on the AVA-Reviews dataset. The notations of IAC, V2L, and MT baseline denote the models for comparision, namely, the image aesthetic classification model, vision-to-language model, and multi-task baseline, respectively. From Table \ref{tab:evalution}, we can observe that (1) the MT baseline achieves inferior results, suggesting that changing the weights of the CNN components has a negative impact; (2) Compared to IAC, V2L, and MT baseline, the proposed Model-\uppercase\expandafter{\romannumeral1} achieves roughly 4.0 $\%$ average improvement and 2.0 $\%$ improvement with regard to image aesthetic prediction and vision-to-language generation, respectively, which suggests that the shared aesthetically semantic layer at a high level can improve the performance of aesthetic prediction as well as that of vision-to-language. (3) Compared to Model-\uppercase\expandafter{\romannumeral1}, the proposed Model-\uppercase\expandafter{\romannumeral2} further improves the performance of aesthetic prediction as well as vision-to-language. It suggests that the combination of task-specific information and task-shared information can further augment the solution with the training over the multi-task architecture.

\begin{figure*}[ht]
   \begin{minipage}{0.20\linewidth}
    \center
    \includegraphics[width=1.0\linewidth]{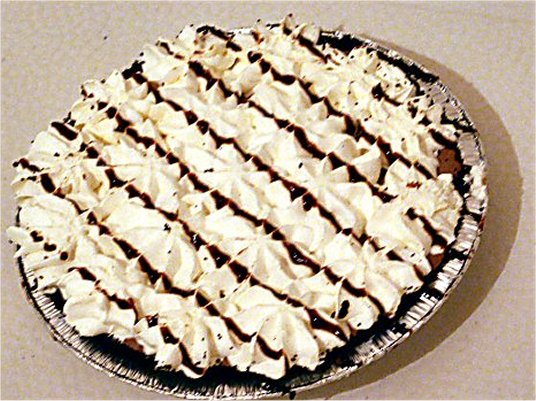}
   \end{minipage}%
   \hfill
   \begin{minipage}{0.29\linewidth}
     \textbf{Ground-truth aesthetic score:} 3.4 (Low-aesthetic category) \\
     \textbf{Ground-truth comments:} Focus is too soft here. Needs to be sharper.\\
                          It's a little blurry \\
     \textbf{Prediction}: Low-aesthetic category\\
     \textbf{Generated comments:} This would have been better if it was more focus.
   \end{minipage}%
   \hfill
   \begin{minipage}{0.20\linewidth}
    \center
    \includegraphics[width=1.0\linewidth]{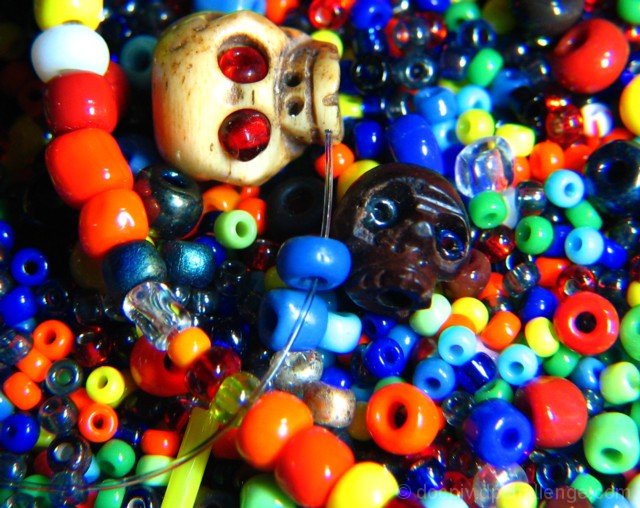}
   \end{minipage}
   \hfill
   \begin{minipage}{0.29\linewidth}
     \textbf{Ground-truth aesthetic score:} 3.7 (Low-aesthetic category) \\
     \textbf{Ground-truth comments:} Too saturated. Not well focused. \\
     \textbf{Prediction:} Low-aesthetic category\\
     \textbf{Generated comments:} The background is a little distracting.
   \end{minipage}

   \vfill

   \begin{minipage}{0.20\linewidth}
    \center
    \includegraphics[width=1.0\linewidth]{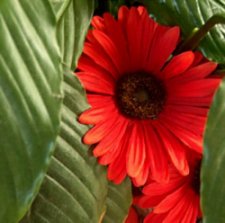}
   \end{minipage}%
   \hfill
   \begin{minipage}{0.29\linewidth}
     \textbf{Ground-truth aesthetic score:} 4.2 (Low-aesthetic category) \\
     \textbf{Ground-truth comments:} Would help if the picture was larger. \\
     \textbf{Prediction}: Low-aesthetic category\\
     \textbf{Generated comments:} The image is too small.
   \end{minipage}%
   \hfill
   \begin{minipage}{0.20\linewidth}
    \center
    \includegraphics[width=1.0\linewidth]{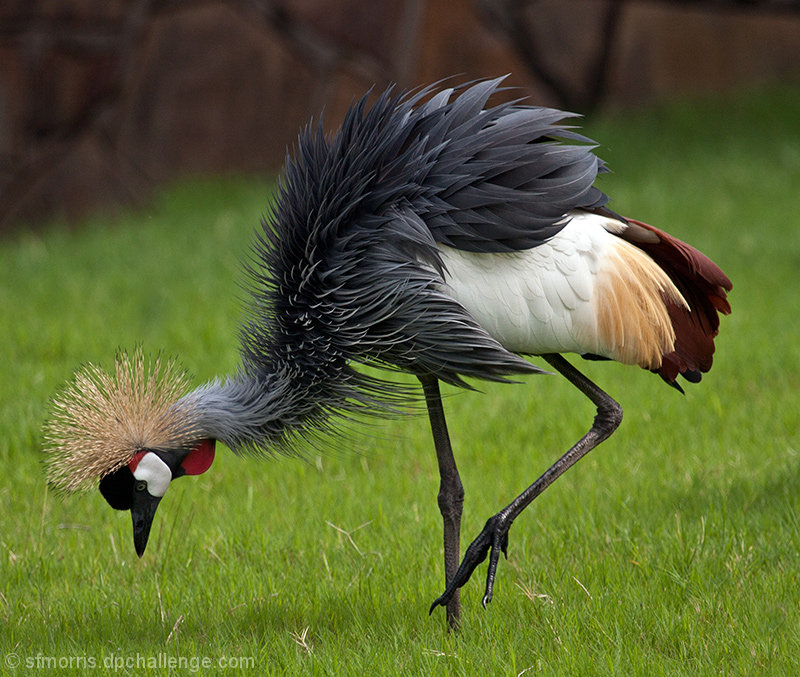}
   \end{minipage}
   \hfill
   \begin{minipage}{0.29\linewidth}
     \textbf{Ground-truth aesthetic score:} 6.1 (High-aesthetic category) \\
     \textbf{Ground-truth comments:} Great detail on this fine bird nicely seperated from background A very good image \\
     \textbf{Prediction}: High-aesthetic category\\
     \textbf{Generated comments:} Great detail in the feathers.
   \end{minipage}

   \vfill

   \begin{minipage}{0.20\linewidth}
    \center
    \includegraphics[width=1.0\linewidth, height=1.0\linewidth]{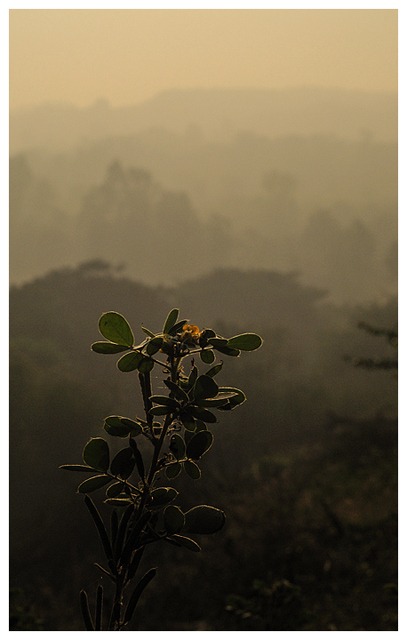}
   \end{minipage}%
   \hfill
   \begin{minipage}{0.29\linewidth}
     \textbf{Ground-truth aesthetic score:} 5.5 (high-aesthetic category) \\
     \textbf{Ground-truth comments:} This is beautiful. The rim lighting on that plant is perfect. \\
     \textbf{Prediction:} High-aesthetic category\\
     \textbf{Generated comments:} The simplicity of this shot.
   \end{minipage}%
   \hfill
   \begin{minipage}{0.20\linewidth}
    \center
    \includegraphics[width=1.0\linewidth]{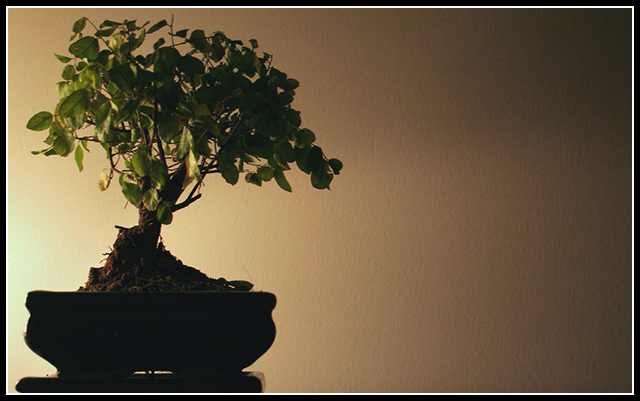}
   \end{minipage}
   \hfill
   \begin{minipage}{0.29\linewidth}
     \textbf{Ground-truth aesthetic score:} 5.6 (high-aesthetic category) \\
     \textbf{Ground-truth comments:} Love the light but the leaves look a bit soft, the composition is also very nice. \\
     \textbf{Prediction}: High-aesthetic category\\
     \textbf{Generated comments:} The lighting is a bit harsh.
   \end{minipage}

   \vfill

   \begin{minipage}{0.20\linewidth}
    \center
    \includegraphics[width=1.0\linewidth]{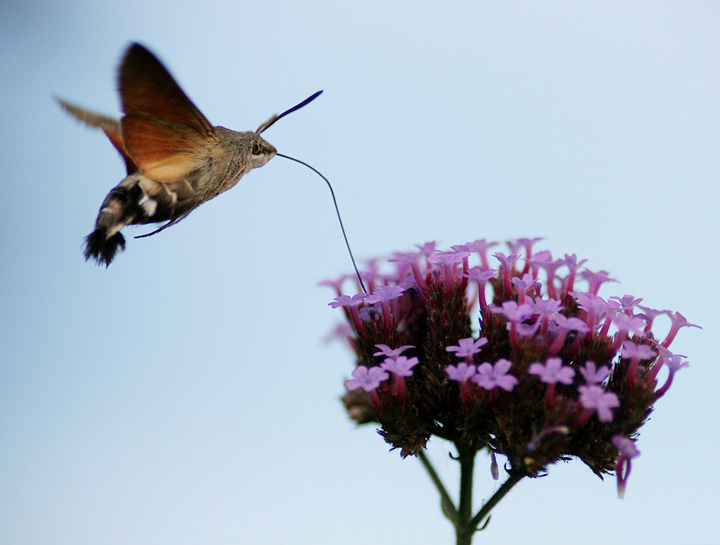}
   \end{minipage}%
   \hfill
   \begin{minipage}{0.29\linewidth}
    \textbf{Ground-truth aesthetic score:} 5.9 (high-aesthetic category) \\
     \textbf{Ground-truth comments:} Tricky shot to nail. Fantastic creatures. \\
     \textbf{Prediction}: High-aesthetic category\\
     \textbf{Generated comments:} Nice capture with the composition and the colors.
   \end{minipage}%
   \hfill
   \begin{minipage}{0.20\linewidth}
    \center
    \includegraphics[width=1.0\linewidth]{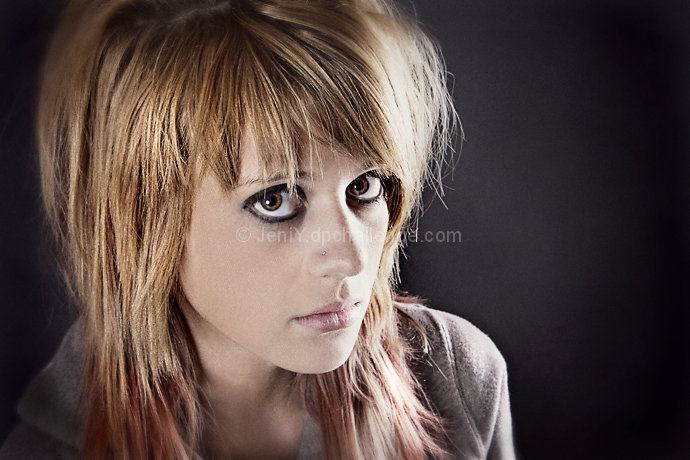}
   \end{minipage}
   \hfill
   \begin{minipage}{0.29\linewidth}
    \textbf{Ground-truth aesthetic score:} 6.08 (high-aesthetic category) \\
     \textbf{Ground-truth comments:} Really good expression captured. \\
      Beautiful huge eyes and great expression and pose.\\
     \textbf{Prediction}: High-aesthetic category\\
     \textbf{Generated comments:} Very nice portrait with the composition.
   \end{minipage}
%
%

   \caption{Typical examples generated by the proposed models.}
   \label{fig:typicalexamples}
 \end{figure*}

 Some representative examples resulting from the proposed models are illustrated in Fig. \ref{fig:typicalexamples}. We can observe that the proposed models can not only achieve satisfactory image aesthetic prediction, but also generate reviews consistent with human cognition. Such examples show that the comments turned out from the proposed models allow the insights into the principles of image aesthetics in broad-spectrum contexts. The proposed models can generate the reviews like ``soft focus'', ``distracting background'', and ``too small'', to suggest why the images obtain low-aesthetic scores. On the other hand, our model also yields the phrases like ``great details'', ``harsh lighting'', and ``very nice portrait'' to give the reasons of assigning high-aesthetic scores. Compared to the ground-truth comments, we see that the proposed models can learn the insights related to aesthetics from the training data.

%

\section{Conclusion}

 In this paper, we investigate into the problem whether computer vision systems have the ability to perform image aesthetic prediction as well as generate reviews to explain why the image of interest leads to a plausible rating score as human cognition. For this sake, we collect the AVA-Reviews dataset to do this research. Specifically, we propose two end-to-end trainable neural models based on CNN plus RNNs architecture, namely, Neural Aesthetic Image Reviewer. By incorporating shared aesthetically semantic layers and task-specific embedding layers at a high level for multi-task learning, the proposed models improve the performance of both tasks. Indeed, the proposed models can promise both image aesthetic classification and generation of natural-language comments simultaneously to aid human-machine cognition.

{\small
\bibliographystyle{ieee}
\bibliography{egbib}
}

\end{document}